\documentclass[10pt,twocolumn,letterpaper]{article}
\usepackage{iccv}
\usepackage{times}
\usepackage{epsfig}
\usepackage{graphicx}
\usepackage{amsmath}
\usepackage{amssymb}
\usepackage[pagebackref=true,breaklinks=true,letterpaper=true,colorlinks,bookmarks=false]{hyperref}
\usepackage{booktabs}
\usepackage{makecell}
\usepackage{multirow}
\usepackage{subcaption}

\usepackage[nocompress]{cite} %
\usepackage[numbers,sort,compress]{natbib}

\usepackage{iccv}
\usepackage{times}
\usepackage{epsfig}
\usepackage{graphicx}
\usepackage{amsmath}
\usepackage{amssymb}

\usepackage[breaklinks=true,bookmarks=false]{hyperref}

\iccvfinalcopy %

\ificcvfinal\fi

\newcommand{\bn}{\mathbf{n}}

\newcommand{\figref}[1]{Fig.~\ref{#1}}
\newcommand{\secref}[1]{Section~\ref{#1}}

\newcommand{\tabref}[1]{Table~\ref{#1}}
\def\sota{SotA }

\makeatletter
\DeclareRobustCommand\onedot{\futurelet\@let@token\@onedot}
\def\@onedot{\ifx\@let@token.\else.\null\fi\xspace}

\makeatother

\newcommand{\boldparagraph}[1]{\vspace{0.2cm}\noindent{\bf #1:}}

\newif\ifshowcomments

\ifshowcomments
    \newcommand{\oh}[1]{ \noindent {\color{red} {\bf OH:} {#1}} }
    \newcommand{\ag}[1]{ \noindent {\color{red} {\bf AG:} {#1}} }
    \newcommand{\mjb}[1]{ \noindent {\color{red} {\bf MJB:} {#1}} }
    \newcommand{\jy}[1]{ \noindent {\color{red} {\bf JY:} {#1}} }
    \newcommand{\xc}[1]{ \noindent {\color{red} {\bf XC:} {#1}} }
    \newcommand{\zj}[1]{ \noindent {\color{red} {\bf Zijian:} {#1}} }
    \newcommand{\todo}[1]{ \noindent {\color{blue} {\bf TODO:} {#1}} }
 \else
    \newcommand{\oh}[1]{\unskip}
    \newcommand{\ag}[1]{\unskip}
    \newcommand{\mjb}[1]{\unskip}
    \newcommand{\jy}[1]{\unskip}
    \newcommand{\xc}[1]{\unskip}
    \newcommand{\zj}[1]{\unskip}
    \newcommand{\todo}[1]{\unskip}
\fi

\newcommand{\point}{\mathbf{x}}
\newcommand{\predcolor}{\mathbf{c}}

\newcommand{\latent}{\mathbf{z}}

\newcommand{\cfeature}{\mathbf{f}}
\newcommand{\ray}{\mathbf{r}}
\newcommand{\camo}{\mathbf{o}}
\newcommand{\viewd}{\mathbf{v}}
\newcommand{\normal}{\mathbf{n}}

\newcommand{\pose}{\mathbf{p}}

\newcommand{\loss}{\mathcal{L}}

\newcommand{\supp}{Sup.~Mat}

\newcommand{\den}{\sigma}
\newcommand{\fidn}{\text{FID}_{\text{normal}}}
\newcommand{\fidf}{\text{FID}_{\text{face}}}
\newcommand{\fidi}{\text{FID}_{\text{image}}}

\newcommand{\R}[1]{\mathbb{R}^{#1}}

\begin{document}

\title{AG3D: Learning to Generate 3D Avatars from 2D Image Collections}

\author{\vspace{.05cm} Zijian Dong$^{1,2*}$ \quad Xu Chen$^{1,3*}$ \quad Jinlong Yang$^{3}$ \quad Michael J. Black$^{3}$ \quad Otmar Hilliges$^{1}$ \quad Andreas Geiger$^{2}$ \\
\vspace{.05cm}$^1$ETH Z{\"u}rich, Department of Computer Science \quad $^2$University of Tübingen \\ $^3$Max Planck Institute for Intelligent Systems, T{\"u}bingen
}

\twocolumn[{%
\renewcommand\twocolumn[1][]{#1}%
\maketitle
\vspace{-4em}
\begin{center}
    \includegraphics[width=\textwidth,trim=0 0 0 0, clip]{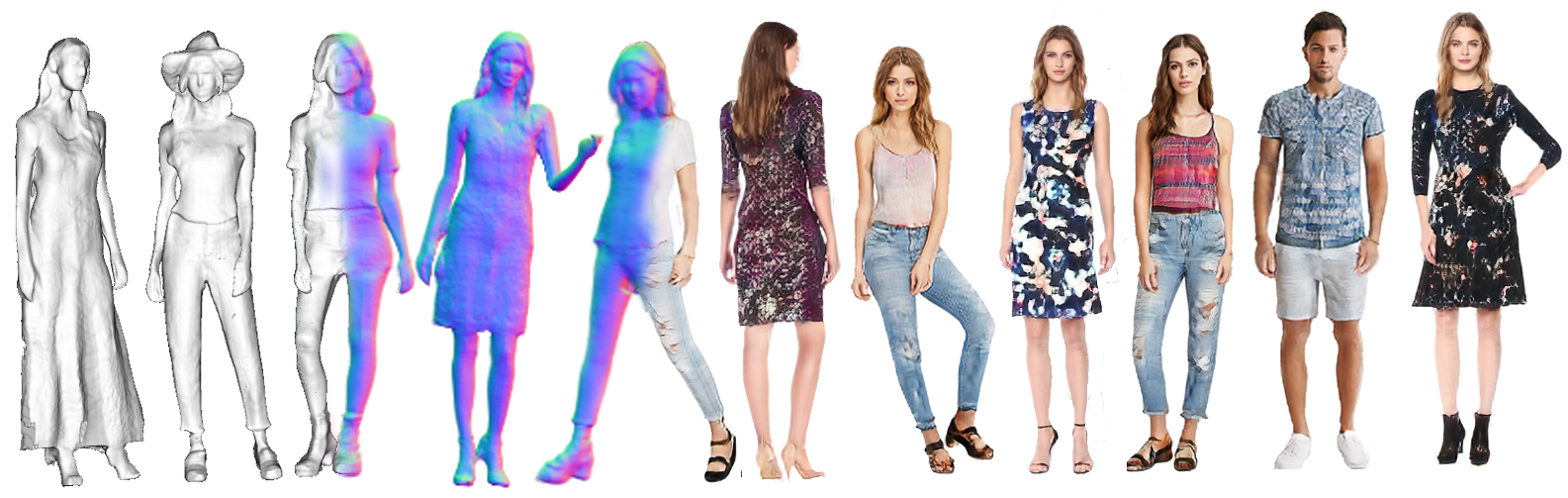}
    \vspace{-2em}
    \captionof{figure}{\textbf{Sampled 3D Human Appearance and Shape.} Our generative model is learned from an unstructured 2D image collection, yet synthesizes novel 3D humans with high-quality appearance and geometry, different identities and clothing styles including loose clothing such as dresses and skirts. Moreover, our generated 3D humans can be easily animated. }
    \label{fig:teaser}
\end{center}
}
]
\ificcvfinal\thispagestyle{empty}\fi

\def\thefootnote{*}\footnotetext{Equal contribution}
\begin{abstract}
\vspace{-1em}
While progress in 2D generative models of human appearance has been rapid, many applications require 3D avatars that can be animated and rendered. Unfortunately, most existing methods for learning generative models of 3D humans with diverse shape and appearance require 3D training data, which is limited and expensive to acquire. 
The key to progress is hence to learn generative models of 3D avatars from abundant unstructured 2D image collections. 
However, learning realistic and complete 3D appearance and geometry in this under-constrained setting remains challenging, especially in the presence of loose clothing such as dresses.
In this paper, we propose a new adversarial generative model of realistic 3D people from 2D images. 
Our method captures shape and deformation of the body and loose clothing by adopting a holistic 3D generator and integrating an efficient and flexible articulation module. 
To improve realism, we train our model using multiple discriminators while also integrating geometric cues in the form of predicted 2D normal maps. 
We experimentally find that our method outperforms previous 3D- and articulation-aware methods in terms of geometry and appearance. 
We validate the effectiveness of our model and the importance of each component via systematic ablation studies.

\end{abstract}

\section{Introduction}
Generative models, like GANs~\cite{goodfellow2020generative}, can be trained from large image collections, to produce  photo-realistic images of objects~\cite{brock2018large, karras2017progressive, karras2019style, Karras2019stylegan2} and even clothed humans~\cite{fu2022styleganhuman, albahar2021pose, lewis2021tryongan, sarkar2021style,lassner2017generative,grigorev2021stylepeople}. 
The output, however, is only a 2D image and many applications require 
diverse, high-quality, virtual 3D avatars, with the ability to control poses and camera viewpoints, while ensuring 3D consistency. 
To enable the generation of 3D avatars, the research community has been studying generative models that can automatically produce 3D shapes of humans and/or clothing based on input parameters such as body pose and shape~\cite{loper2015smpl, chen2022gdna, corona2021smplicit, palafox2021npms}. Despite rapid progress, most existing methods do not yet consider texture and require accurate and clean 3D scans of humans for training, which are expensive to acquire and hence limited in quantity and diversity. 
In this paper, we develop a method that learns a generative model of 3D humans with texture from only a set of unstructured 2D images of various people in different poses wearing diverse clothing; that is, we learn a generative 3D human model from data that is ubiquitous on the Internet. 

Learning to generate 3D shapes and textures of articulated humans from such unstructured image data  is a highly under-constrained problem, as each training instance has a different shape and appearance and is observed only once from a particular viewpoint and in a particular pose. 
Recent progress in 3D-aware GANs~\cite{chan2022efficient,or2022stylesdf,gu2021stylenerf} shows impressive results in learning 3D geometry and appearance of rigid objects from 2D image collections. However, since humans are highly articulated and have more degrees of freedom to model, such methods struggle to generate realistic humans. 
By modeling articulation, recent work~\cite{noguchi2022unsupervised,
bergman2022generative} demonstrates the feasibility of learning articulated humans from image collections, allowing the generation of human shapes and images in desired poses, but only in limited quality and resolution. %
Recently, EVA3D~\cite{hong2022eva3d} achieves higher resolution by representing humans as a composition of multiple parts, each of which are generated by a small network. 
However, there is still a noticeable gap between the generated and real humans in terms of appearance and, in particular, geometry.
Additionally, the compositional design precludes modeling loose clothing that is not associated with a single body part, such as dresses shown in Fig.~\ref{fig:result_compare:dress}. \todo{more precise, one experiment for each limitation, should we ref the figure/exp}

\oh{the following two paragraphs do a good job in describing the individual components of your contribution. I would try to pre-face this with a summary on the conceptual level. I.e., what is your philosophy? Use a monolithic approach for the generator to model body and loose clothing, use a modular discriminator to increase fidelity of small regions and normals.} \todo{we try to merge this comment, but still can't find a proper way to motivate it naturally.  }

In this paper, we contribute a new method for learning 3D human generation from 2D image collections, which yields state-of-the-art image and geometry quality and naturally models loose clothing. Instead of representing humans with separate body parts as in EVA3D~\cite{hong2022eva3d}, we adopt a simple monolithic approach that is able to model the human body as well as loose clothing, while adding multiple discriminators that increase the fidelity of perceptually important regions and  improve geometric details. 

\boldparagraph{Holistic 3D Generation and Deformation} 
To achieve the goal of high image quality while flexibly handling loose clothing, we propose a novel generator design. We model 3D humans holistically in a canonical space using a monolithic 3D generator and an efficient tri-plane representation~\cite{chan2022efficient}. 
An important aspect in attaining high-quality images is to enable fast volume rendering. 
To this end, we adapt the efficient articulation module, Fast-SNARF~\cite{chen2022fastsnarf}, to our generative setting and further accelerate rendering via empty-space skipping, informed by a coarse human body prior. 
Our articulation module is more flexible than prior methods that base deformations of the clothed body on SMPL \cite{loper2015smpl}, enabling it to  faithfully model deformations for points that are far away from the body.

\boldparagraph{Modular 2D Discriminators} We further propose multiple discriminators to improve geometric detail as well as the perceptually-important face region as we found that a single adversarial loss on rendered images is insufficient to recover meaningful 3D geometry in such a highly under-constrained setting. 
Motivated by the recent success of methods~\cite{jiang2022selfrecon,Yu2022MonoSDF} that exploit monocular normal cues \cite{saito2020pifuhd,xiu2022icon} for the task of 3D reconstruction,
we explore the utility of normal information for guiding 3D geometry in  the generative setting. 
More specifically, we discriminate normal maps rendered from our generative 3D model against 2D normal maps obtained from off-the-shelf monocular estimators \cite{saito2020pifuhd} applied to 2D images of human subjects.
We demonstrate that this additional normal supervision serves as useful and complementary guidance, significantly improving the quality of the generated 3D shapes. 
Furthermore, we apply separate face discriminators on both the image and normal branch to encourage more realistic face generation.

We experimentally find that our method outperforms previous 3D- and articulation-aware methods by a large margin in terms of both geometry and texture quality, quantitatively (\tabref{table:main}), qualitatively (\figref{fig:result_compare}) and through a perceptual study (\figref{fig:user_study}). In summary, we contribute (i) a generative model of articulated 3D humans with \sota appearance and geometry, (ii) a new generator that is efficient and can generate and deform loose clothing, and (iii) several, specialized discriminators that significantly improve visual and geometric fidelity. We will release code and models.

\section{Related Work}

\noindent{\bf 3D-aware Generative Adversarial Networks:}
Generative adversarial networks (GANs)~\cite{goodfellow2020generative} achieve photorealistic image generation~\cite{Karras2019stylegan2,karras2017progressive,brock2018large, karras2019style} and show impressive results on the task of 2D human image synthesis~\cite{fu2022styleganhuman, albahar2021pose, lewis2021tryongan, sarkar2021style,lassner2017generative,grigorev2021stylepeople}.
However, these 2D methods cannot guarantee 3D consistency~\cite{ma2017pose, chen2019unpaired,lassner2017generative} and do not provide 3D geometry.  
Several methods extend 2D GANs to 3D by combining them with 3D representations, including 3D voxels \cite{wu2016learning, nguyen2019hologan}, meshes \cite{szabo2019unsupervised, liao2020towards} and point clouds \cite{achlioptas2018learning, li2019pu}. 
Recently, many methods represent 3D objects as neural implicit functions~\cite{mescheder2019occupancy, park2019deepsdf, yariv2021volume, wang2021neus, niemeyer2020differentiable}. Such representations are also used for 3D-aware generative image synthesis~\cite{schwarz2020graf, niemeyer2021giraffe, chan2022efficient, or2022stylesdf,gu2021stylenerf, chan2021pi, schwarzvoxgraf}. StyleSDF~\cite{or2022stylesdf} replaces density with an SDF for better geometry generation and \sota methods like EG3D~\cite{chan2022efficient} introduce a tri-plane representation to improve rendering efficiency. Nevertheless, it is not straightforward to extend these methods to non-rigid articulated objects such as humans. In this paper, we propose a 3D- and articulation-aware generative model for clothed humans.

\boldparagraph{3D Human Models} Parametric 3D human body models~\cite{anguelov2005scape, loper2015smpl, osman2020star,xu2020ghum,joo2018total} are able to synthesize minimally clothed human shapes by deforming a template mesh. 
Extending these mesh models to generate 3D clothing or clothed humans is challenging \cite{ma2020learning}.
In the case of meshes, the geometry is restricted to a fixed mesh topology and large deviations from the template mesh are hard to model.  To overcome this limitation, methods such as SMPLicit~\cite{corona2021smplicit} and gDNA~\cite{chen2022gdna} propose to build a 3D generative model of clothed humans based on implicit surface representations, either by adding an implicit garment to the SMPL body or by learning a multi-subject implicit representation with corresponding skinning weights.  The main problem of all aforementioned approaches, however, is their reliance on 3D ground truth: their training requires a large number of complete and registered 3D scans of clothed humans in different poses, which are typically acquired using expensive 3D body scanners.  Several methods~\cite{dong2022pina, peng2021neural, zhao2022humannerf, wang2022arah, jiang2022neuman, weng2022humannerf, kwon2021neural, Feng2022scarf} combine NeRF with human priors to enable 3D human reconstruction from multi-view data or even monocular videos. Nevertheless, their proposed human representations can only be utilized to represent human
avatars for a single subject, wearing a specific garment.  

Recently, some methods have been proposed to learn generative models of 3D human appearance from a collection of 2D images. ENARF-GAN~\cite{noguchi2022unsupervised} and GNARF~\cite{bergman2022generative} leverage 3D human models to learn a 3D human GAN, but they still fail to produce high-quality human images.
The concurrent work EVA3D~\cite{hong2022eva3d} achieves high-resolution human image generation by introducing a compositional part-based human representation.  However, none of these methods including other concurrent arXiv papers~\cite{zhang2022avatargen,yang20223dhumangan,jiang2022humangen} are able to generate and deform loose clothing, and their geometry typically suffers from noisy artifacts. In contrast, our method generates both high-quality geometry and appearance of diverse 3D-clothed humans even wearing loose clothing, with full control over the pose and appearance. We empirically demonstrate the benefits of our method over the more complex EVA3D model~\cite{hong2022eva3d} in \secref{sec:compare}. A comparison to the recent arXiv papers~\cite{zhang2022avatargen,yang20223dhumangan,jiang2022humangen} is not possible since the models and code have not been released.

\boldparagraph{3D Shape from 2D Normals} 
\todo{whether to add: 3D normals have been widely explored for enhancing geometrical details of 2D representations such as depth~\cite{lahner2018deepwrinkles} maps or UV normal~\cite{wang2020normalgan}, and 3D normal fields~\cite{chen2022gdna}. However, all of these methods require ground-truth shapes and 3D normal supervision, which is difficult to obtain.} 
Several methods predict normals from a single image, for general objects~\cite{eigen2014depth, eigen2015predicting, eftekhar2021omnidata, huang2022good} or clothed humans~\cite{saito2020pifuhd,xiu2022icon}. 
These predicted 2D normal cues can be exploited to guide 3D reconstruction using neural field representations. For instance, MonoSDF~\cite{Yu2022MonoSDF} leverages predicted normals to improve 3D object reconstruction from sparse views. Similarly, SelfRecon~\cite{jiang2022selfrecon} uses a normal reconstruction loss to reconstruct a human avatar from a monocular video. PIFuHD~\cite{saito2020pifuhd} and ICON~\cite{xiu2022icon} predict normal maps as additional input to support single-view 3D human reconstruction. 
In this work, we demonstrate that monocular 2D normal cues are useful for learning a generative 3D model of articulated objects.

\section{Method}

\begin{figure*}
    \centering
    \includegraphics[width=\linewidth, trim=25 0 0 10,clip]{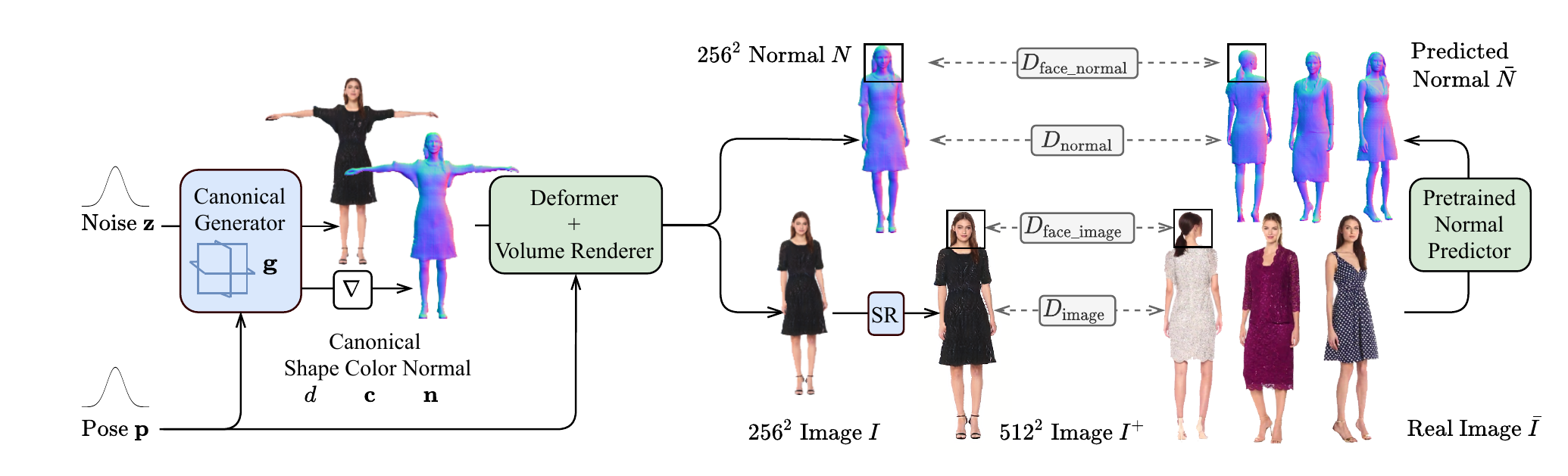}
    \vspace{-2.5em}
    \caption{\textbf{Method Overview.} \emph{Holistic 3D Human Generation:} Given a latent vector $\mathbf{z}$, our method generates human shape $d$ and appearance $\mathbf{c}$ in canonical space. In addition, we compute surface normals $\bn$ via the spatial derivatives of the canonical shape which is represented as an SDF. These canonical representations are then posed into the target body pose $\mathbf{p}$ via a flexible deformer and then rendered from the target viewpoint. The rendered images are further super-resolved by $2\times$. \emph{Adversarial Training:} We optimize the generator and the super-resolution module using multiple discriminators. In addition to an image discriminator operating on the images, we improve geometry by introducing a normal discriminator that compares our rendered normal maps with the normals of real images predicted by an off-the-shelf normal estimator. To further improve the quality of the perceptually important face region, we add normal and image discriminators for the face region.}
    \label{fig:pipeline}
    \vspace{-0.3cm}
\end{figure*}

Given a large 2D image collection, our goal is to learn a generative model of diverse 3D human avatars with realistic appearance and geometry, while enabling control over pose and identity.  An overview of our method is shown in \figref{fig:pipeline}.

In this section, we first introduce an efficient and articulation-aware 3D human generator (\secref{sec:avatar_generator})
which generates the appearance and shape in canonical space and uses a deformation module to warp into posed space via a learned continuous deformation field.
Next, we describe our rendering module that is accelerated by an empty space skipping strategy which leverages the SMPL body prior. 
To enable fast training, we use a super-resolution module to lift feature maps to high-resolution images.

We optimize the generator using a combination of adversarial losses (\secref{sec:gan_training}) and an Eikonal loss~\cite{gropp2020implicit}. 
While prior work uses a single discriminator formulation, we show that employing several, specialized discriminators improves visual fidelity. 
To this end, we define discriminators that reason at the level of the whole body and locally at the face region, respectively.  
We additionally introduce an adversarial normal loss, which significantly improves the quality of the generated geometry.

\subsection{ Holistic 3D Avatar Generator}
\todo{maybe change to:  Holistic  3D Avatar Generator}
\label{sec:avatar_generator}
\label{sec:model}

\boldparagraph{Canonical Generator} 
Given a latent vector $\latent \in \R{n_z}$ and pose parameters $\pose \in \R{n_p}$, our method first generates 3D human appearance and shape in canonical space (see \figref{fig:pipeline}). Here, we leverage pose conditioning to model pose-dependent effects. For efficient rendering, the canonical generator builds on the tri-plane representation proposed in ConvONet~\cite{peng2020convolutional} and EG3D~\cite{chan2022efficient} to model 3D features. These are then decoded by an MLP to predict the canonical shape and appearance in 3D space. 

We represent geometry using a signed distance field (SDF). Since existing fashion datasets are  imbalanced and contain mostly frontal views, learning correct 3D geometry from such datasets is difficult. 
Following ~\cite{hong2022eva3d}, we exploit a human shape prior in the form of canonical SMPL~\cite{loper2015smpl}. 
Specifically, for every query point $\point$ in canonical space, we predict an SDF offset $\Delta d(\point, \latent, \pose)$ from the base shape to model details such as hair and clothing. 
The SDF value $d(\point, \latent, \pose)$ is then calculated as
\begin{align}
d = d(\point, \latent, \pose) = d_\text{SMPL}(\point) + \Delta d(\point, \latent, \pose) ,
\end{align} 
where $d_\text{SMPL}(\point)$ is the signed distance to the canonical SMPL surface. Unlike \cite{hong2022eva3d}, to compute $d_\text{SMPL}(\point) $  efficiently, we represent the SDF as a low-resolution voxel grid ($128\times128\times32$), where the value of every grid point
is the (pre-computed) distance to the SMPL mesh. We then query $d_\text{SMPL}(\point) $ by trilinear interpolating the SDF voxel grid. 

We also compute normals in canonical space. The normal $\normal$ at a certain canonical point $\point$ is computed as the spatial gradient of the signed distance function at that point:
\begin{equation}
\normal = \nabla_{\point}d(\point, \latent, \pose) .
\end{equation}
The canonical appearance is represented by a 3D texture field $\predcolor$. We also predict features $\cfeature$ that are used to guide the super-resolution module (described later).
We denote the entire mapping from 3D point $\point$, latent vector $\latent$ and pose condition $\pose$ to SDF $d$, normal $\normal$, color $\predcolor$ and color features $\cfeature$ in \textit{canonical space} as follows:
\begin{align}
\mathbf{g}: \R{3} \times \R{n_z}  \times \R{n_p}   &\rightarrow \mathbb{R} \times \R{3} \times \R{3} \times \R{n_f} \\
        (\point, \latent, \pose)  &\mapsto (d, \normal, \predcolor, \cfeature). \nonumber
\label{eqn:occ}
\end{align}

\boldparagraph{Deformer}
To enable animation and to learn from posed images, we require the appearance and 3D shape in \textit{posed space}. In the following we denote quantities in posed space as $(\cdot)'$.
Given the bone transformation matrix $\mathbf{B}_i$ for joint $i \in \{1,...,n_b\}$, a canonical point $\point$ is transformed into its deformed version $\point'$ via
\begin{equation}
\label{eq:deform}
\mathbf{x}' = \sum_{i = 1}^{n_b} w_i \, \mathbf{B}_i \, \mathbf{x}
\end{equation}
Here, the canonical LBS weight field $\mathbf{w}: \R{3} \rightarrow \R{n_b}$, with $\point \mapsto (w_{1},...,w_{n_b})$ and $n_b$ the number of joints, weights the influence of each bone's $\mathbf{B}_i$ transformation onto the deformed location $\mathbf{x}'$. This weight field is represented by a low-resolution voxel grid. 
The normal at the deformed point $\point'$ is given by
\begin{equation}
    \normal' = \frac{(\sum_{i = 1}^{n_b}w_i\, \mathbf{R}_i \,)^{-T} \mathbf{n}}{\|(\sum_{i = 1}^{n_b}w_i\, \mathbf{R}_i \,)^{-T} \mathbf{n}\|}
\end{equation}
where $\mathbf{R}_i$ is the rotation component of $\mathbf{B}_i$~\cite{tarini2014accurate}. 

We leverage Fast-SNARF~\cite{chen2022fastsnarf} to efficiently warp points \textit{backwards} from posed space $\point'$ to canonical space $\point$ via efficient iterative root finding~\cite{chen2022fastsnarf}. 
The SDF value $d'$, color $\predcolor'$ and feature $\cfeature'$ at the deformed point are obtained by evaluating the generator at the corresponding $\point$.
In contrast to~\cite{chen2022fastsnarf}, which focuses on reconstruction tasks and learns skinning weights on the fly, we constrain the notoriously difficult adversarial training
by averaging the skinning weights of the nearest vertices on the canonical SMPL mesh.

\boldparagraph{Volume Renderer}
To render a pixel, we follow~\cite{mildenhall2021nerf} and cast ray $\ray'$ from the camera center $\camo'$ along its view direction $\viewd'$. We use two-pass importance sampling of $M$ points in posed space  $\point_i' = \camo' + t_{i}\viewd'$ and predict their SDF values $d_{i}'$, colors $\predcolor_i'$, color features $\cfeature_{i}'$ and normals $\normal_{i}'$.
We convert SDF values $d_{i}'$ to densities $\den_i'$ via the method of StyleSDF~\cite{or2022stylesdf}.
\begin{figure*}
     \centering
     \begin{subfigure}{0.25\textwidth}
         \centering \includegraphics[width=\textwidth]{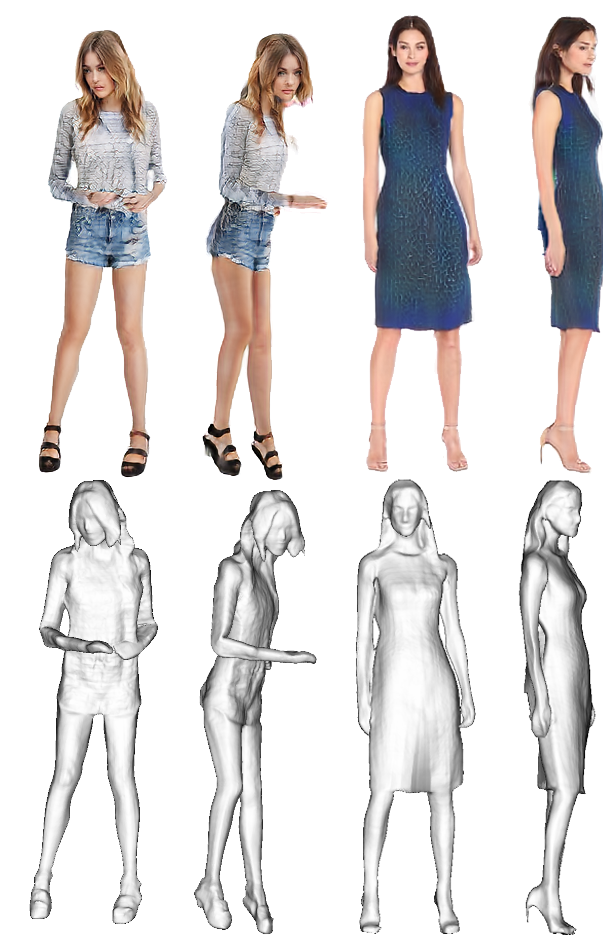}
         \caption{View Control}
         \label{fig:result_sample:view}
     \end{subfigure}
     \hfill
      \begin{subfigure}{0.25\textwidth}
         \centering
         \includegraphics[width=\textwidth]{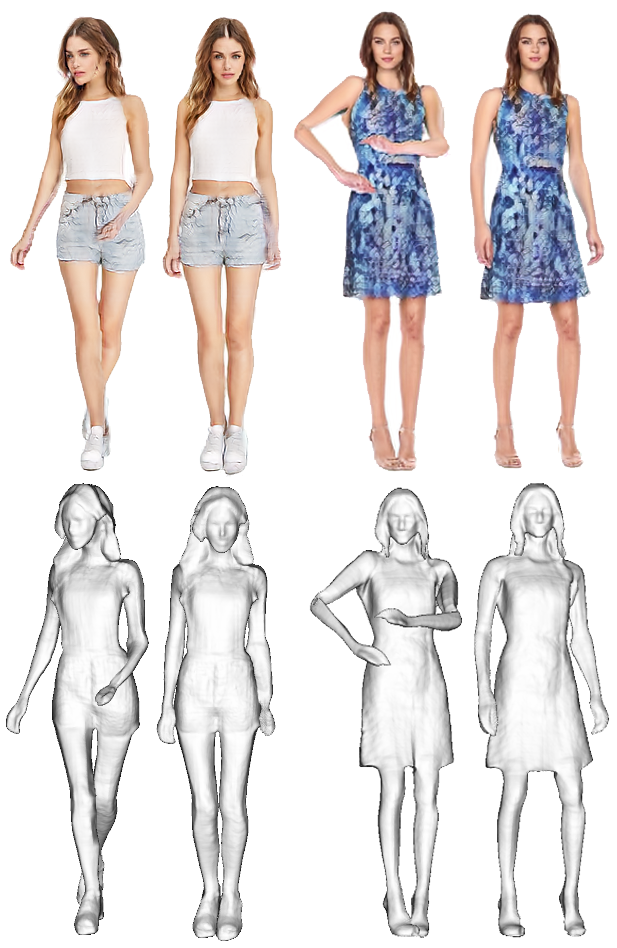}
         \caption{Pose Control}
         \label{fig:result_sample:pose}
     \end{subfigure}
     \hfill
     \begin{subfigure}{0.48\textwidth}
         \centering
         \includegraphics[width=\textwidth]{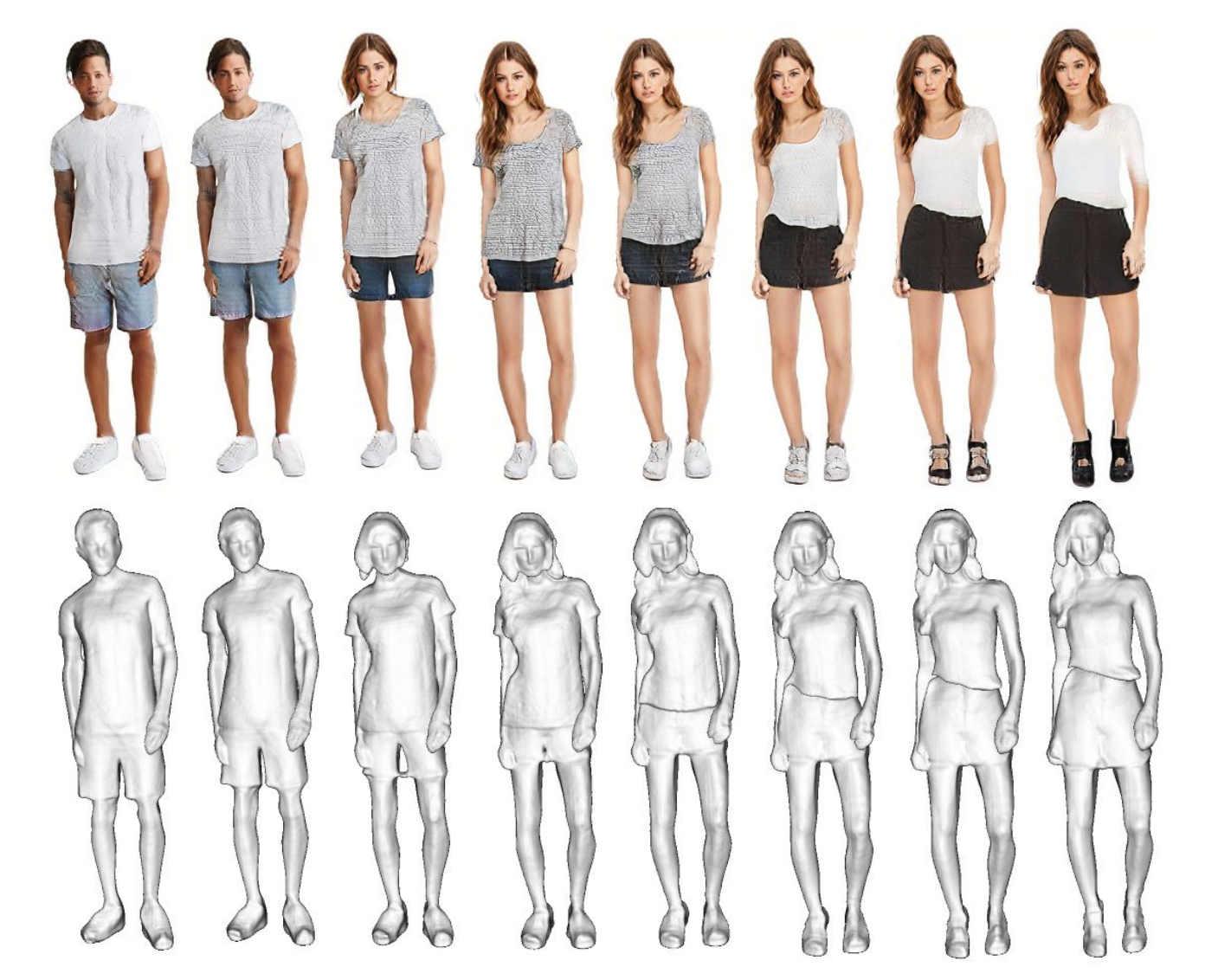}
         \caption{Interpolation}
         \label{fig:result_sample:interp}
     \end{subfigure}
     \vspace{-0.1in}
     \caption{\textbf{Qualitative Results: 3D Human Generation.} We generate 3D human appearance and shape, and render the resulting 3D representations using different body poses and from different viewpoints. In addition, we show virtual people generated by interpolating between latent codes. Overall, our synthesized humans exhibit reasonable appearance and geometric quality, remain consistent across different poses and views, and smoothly interpolate when varying the latent code $\mathbf{z}$.
    }
\label{fig:result_sample}
\vspace{-0.3cm}
\end{figure*}

The color of each pixel in the rendered image $I$ is computed via numerical integration~\cite{mildenhall2021nerf}:
\begin{equation}
    I(\ray) = \sum_{i = 1}^{M} \alpha_i \prod_{i<j} (1-\alpha_j) \predcolor_{i}' \quad \alpha_i = 1-\text{exp}(\den_{i}' \delta_i)
\end{equation}
where $\delta_i$ is the distance between samples. 3D normals $N(\ray)$ and feature vectors $F(\ray)$ are rendered accordingly.
To accelerate rendering and to reduce memory, we take advantage of the geometric prior of the SMPL model and define the region within a predefined distance threshold to the SMPL surface as the occupied region. For points sampled outside of this region, we set the density to zero.

 \boldparagraph{Super Resolution} Although the SMPL-guided volume rendering is more efficient than previous approaches, it is still slow and requires a large amount of memory to render at high resolution. Therefore, we perform volume rendering at a sufficient resolution ($256^2$ pixels) to guarantee good rendering of the normal image $N$ and rely on a super-resolution module~\cite{chan2022efficient} to upsample the image feature map $F$ and color $I$ to the final image $I^{+}$ of size $512^2$ pixels. 
 
\todo{what is new, what has been done before: currently, for the existing method, we cite them and use following..., others are new: normal part, SMPL-guided rendering, SMPL template SDF part.}

\subsection{Training}
\label{sec:gan_training}
We train our model on a large dataset of 2D images using adversarial training,
leveraging a combination of multiple discriminators and an Eikonal loss.

\boldparagraph{Image Discriminator}
The first discriminator $D_{\text{image}}$ compares full images generated by our method to real images. Following EG3D~\cite{chan2022efficient}, we apply the discriminator at both resolutions: We upsample our low resolution rendering $I$, concatenate it with the super-resolved image $I^+$, and feed it to a StyleGANv2~\cite{karras2019style} discriminator. For real images $\bar{I}$, we downsample and re-upsample them, and concatenate the results with the original image as input to the discriminator.

\boldparagraph{Face Discriminator} We observe that the generated face region suffers from artifacts due to the low resolution of faces within the full-body image. Motivated by 2D human GANs~\cite{grigorev2021stylepeople}, we add a small face discriminator $D_{\text{face\_image}}$. Based on estimated SMPL head keypoints, we crop the head regions of our high resolution output $I^+$ and real data $\bar{I}$ and feed them into the discriminator for comparison. 

\boldparagraph{Normal Discriminator} A central goal of our work is to attain geometrically correct 3D avatars. 
To achieve this, we propose to use geometric cues present in 2D normal maps to guide the adversarial learning towards meaningful 3D geometry. To this end, we use an additional normal discriminator $D_{\text{normal}}$. This normal discriminator compares the predicted 2D normal maps $N$ to 2D normal maps $\bar{N}$ of real images $\bar{I}$ predicted by the 2D normal estimator from PIFuHD~\cite{saito2020pifuhd}. 
Analogously to the image branch, we use an additional discriminator $D_{\text{face\_normal}}$ to further enhance the geometric fidelity of the generated faces.
We refer the reader to the Sup.~Mat.~for implementation details.

\boldparagraph{Eikonal Loss} 
To regularize the learned SDFs, we apply an Eikonal loss~\cite{gropp2020implicit} $\loss_{\text{eik}} =  \mathbb{E}_{\mathbf{x}_{i}}\left(\|\nabla (\Delta d(\mathbf{x}_{i}))\| -1\right)^{2}$ to the canonical correspondences $\{\point_{i}\}$ of sampled points $\{\point'_{i}\}$.

\boldparagraph{Training}
We train our generator and discriminators jointly using the non-saturating GAN objective with R1-regularization~\cite{mescheder2018training} and an Eikonal loss. Please refer to the \supp. for more training details. 

\section{Experiments}
\label{sec:experiment}

\begin{table*}
\centering
\begin{tabular}{l|ccc|ccc}
    \Xhline{1pt}
    \multirow{2}{*}{Method} & \multicolumn{3}{c|}{{DeepFashion}}  & \multicolumn{3}{c}{{UBCFashion}} \\
    & FID$\downarrow$ &$\text{FID}_\text{normal}\downarrow$ &  $\text{FID}_\text{face}\downarrow$ 
    & FID$\downarrow$ & $\text{FID}_\text{normal}\downarrow$ & $\text{FID}_\text{face}\downarrow$  \\
    \hline\hline
    EG3D       & $26.38^{*}$ &-&-      & $23.95^{*}$ &-&-      \\
    StyleSDF   & $92.40^{*}$ &-&-  & $18.52^{*}$ &-&-      \\
    ENARF-GAN & $77.03^{*}$ & -&-&- &-&-  \\
    EVA3D     & $15.91^{*}$ &- &- & $12.61^{*}$ &-&- \\
    \hline
    EVA3D (public)     & 20.45 &30.81 &17.21 & 19.81  &49.29&54.42 \\
    Ours       & \textbf{10.93}  & \textbf{20.38} & \textbf{14.79}     & \textbf{11.04}  &\textbf{18.79}&\textbf{15.83}     \\
    \Xhline{1pt}
\end{tabular}
\caption{\textbf{Quantitative Comparison with \sota Methods}. We evaluate FID of full images, cropped face images and normal maps generated by our method and the \sota method EVA3D (public)~\cite{hong2022eva3d} using their released trained models. For reference, we also report quantitative results from the EVA3D paper above the separation line marked by *.
}

\label{table:main}
\vspace{-0.3cm}
\end{table*}
In our experiments, we first demonstrate the quality of generated samples and then compare our method to other \sota baselines. In addition, we provide an ablation study to investigate the importance of each component in our model.
\begin{figure}
    \centering
    \includegraphics[width=\linewidth]{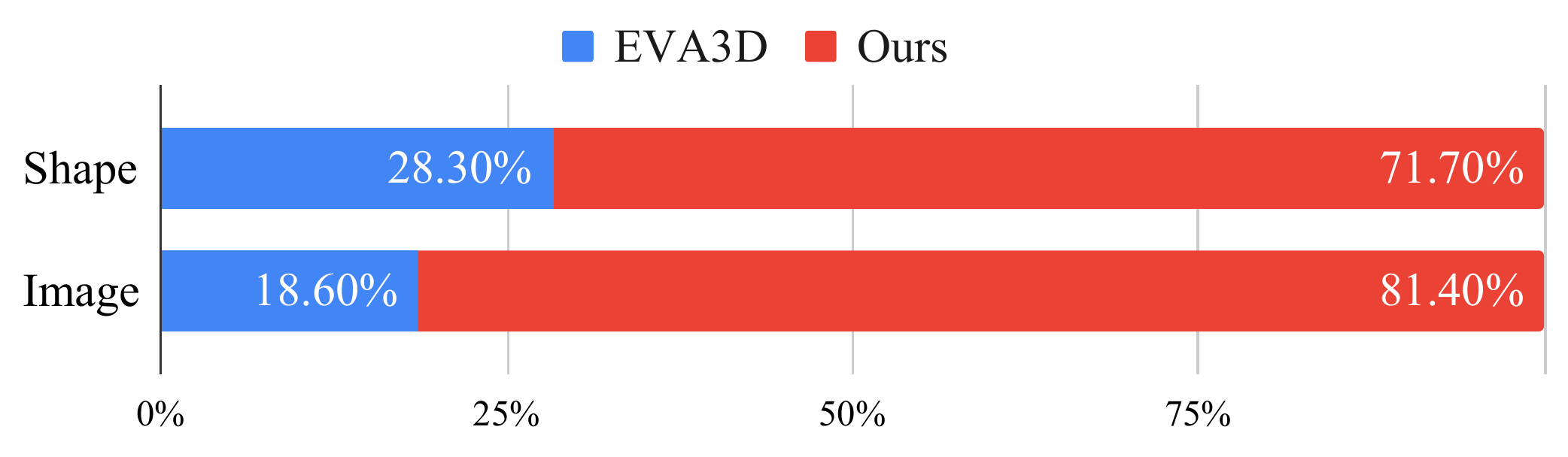}
    \caption{\textbf{User Preference}. We conduct a perceptual study with approximately 4000 samples and report how often participants preferred shapes and images generated by our method or those generated by  EVA3D~\cite{hong2022eva3d}.}
    \label{fig:user_study}
    \vspace{-0.3cm}
\end{figure}

\boldparagraph{Datasets}\footnote{
Disclaimer: Standard fashion datasets lack diversity, see \secref{sec:limitations}.
}
\textit{DeepFashion}~\cite{liu2016deepfashion}
contains an unstructured collection of fashion images of different subjects wearing various types of clothing. We use the curated subset with 8k images from \cite{hong2022eva3d} as our training data.
\textit{UBCFashion}~\cite{zablotskaia2019dwnet} contains 500 sequences of fashion videos with subjects wearing loose clothing such as skirts. Following EVA3D~\cite{hong2022eva3d}, we treat these videos as individual images without assuming temporal information. Pre-processing details can be found in the \supp.

\boldparagraph{Metrics}
We measure the diversity and quality of generated images using the \textit{Fr\'echet Inception Distance (FID)} between 50k generated images and available real images, denoted by $\text{FID}_\text{image}$. To measure the generated face quality, we report an additional FID specifically for the face region, denoted by $\text{FID}_\text{face}$. Furthermore, we evaluate the quality of the synthesized geometry by computing the FID  between our rendered normals and pseudo-GT normal maps predicted by \cite{saito2020pifuhd} ($\text{FID}_\text{normal}$).  We use an inception network~\cite{szegedy2015inception} pre-trained on ImageNet~\cite{deng2009imagenet} for all FID computation.
In addition, we conduct a \textit{Perceptual User Study} among 50 participants with 4000 samples and report how often participants preferred a particular method over ours.

\boldparagraph{Baselines}
We compare our method to four baseline methods: EG3D~\cite{chan2022efficient}, StyleSDF~\cite{or2022stylesdf}, ENARF-GAN~\cite{noguchi2022unsupervised} and EVA3D~\cite{hong2022eva3d}. EG3D and StyleSDF are \sota, %
3D-aware, generative models of rigid objects. For comparison, these two methods are trained on the aforementioned human datasets. Since these two methods do not model articulation, they have to learn it implicitly. 
ENARF-GAN and EVA3D additionally model articulation for 3D human generation. The quantitative FID results of all baseline methods are directly taken from the experiment in EVA3D~\cite{chan2022efficient}. In addition, we evaluate $\text{FID}_\text{face}$ and  $\text{FID}_\text{normal}$ on EVA3D based on their released code and trained model weights.

\subsection{Quality of 3D Human Generation}
We show our qualitative results in \figref{fig:result_sample}. More results can be found in the \supp. Overall, our method generates realistic human images with faithful details such as clothing patterns, face and hair, and meaningful 3D geometry even with fine structures such as hair and shoe heels. Our method further enables control over the generation as follows.

\boldparagraph{View Control}
As shown in \figref{fig:result_sample:view}, by learning humans in 3D space, our method can generate 3D-consistent high-quality images and geometry from varied viewpoints.

\boldparagraph{Pose Control} The generated 3D humans can also be reposed into unseen poses as shown in \figref{fig:result_sample:pose}. The images and geometry in different poses are consistent due to the explicit model of human articulation. 

\begin{figure*}
     \centering
     \begin{subfigure}{0.468\textwidth}
         \centering \frame{\includegraphics[width=\textwidth]{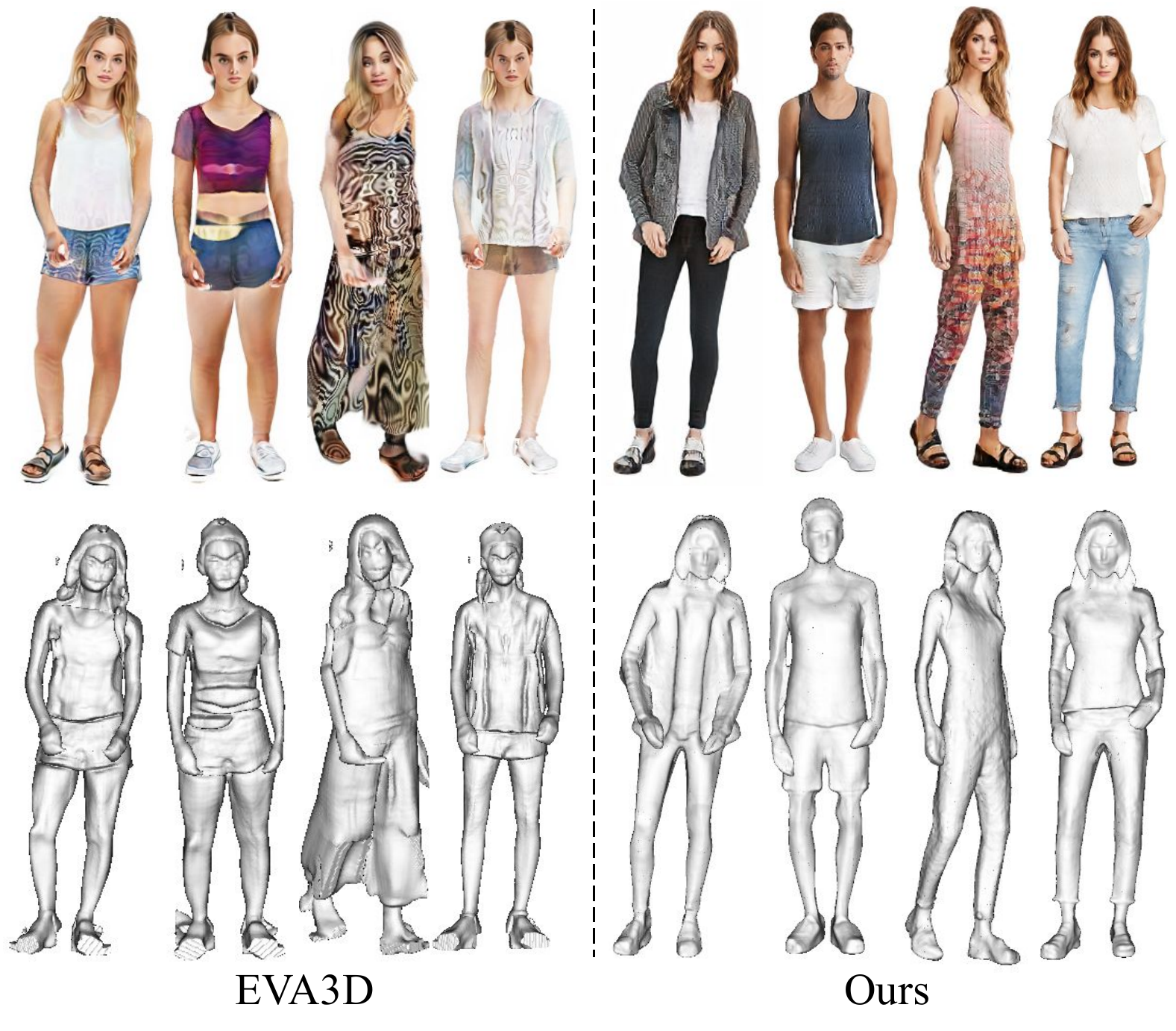}}
         \caption{Overall Quality}
         \label{fig:result_compare:overall}
     \end{subfigure}
     \hfill
      \begin{subfigure}{0.255\textwidth}
         \centering
         \frame{\includegraphics[width=\textwidth]{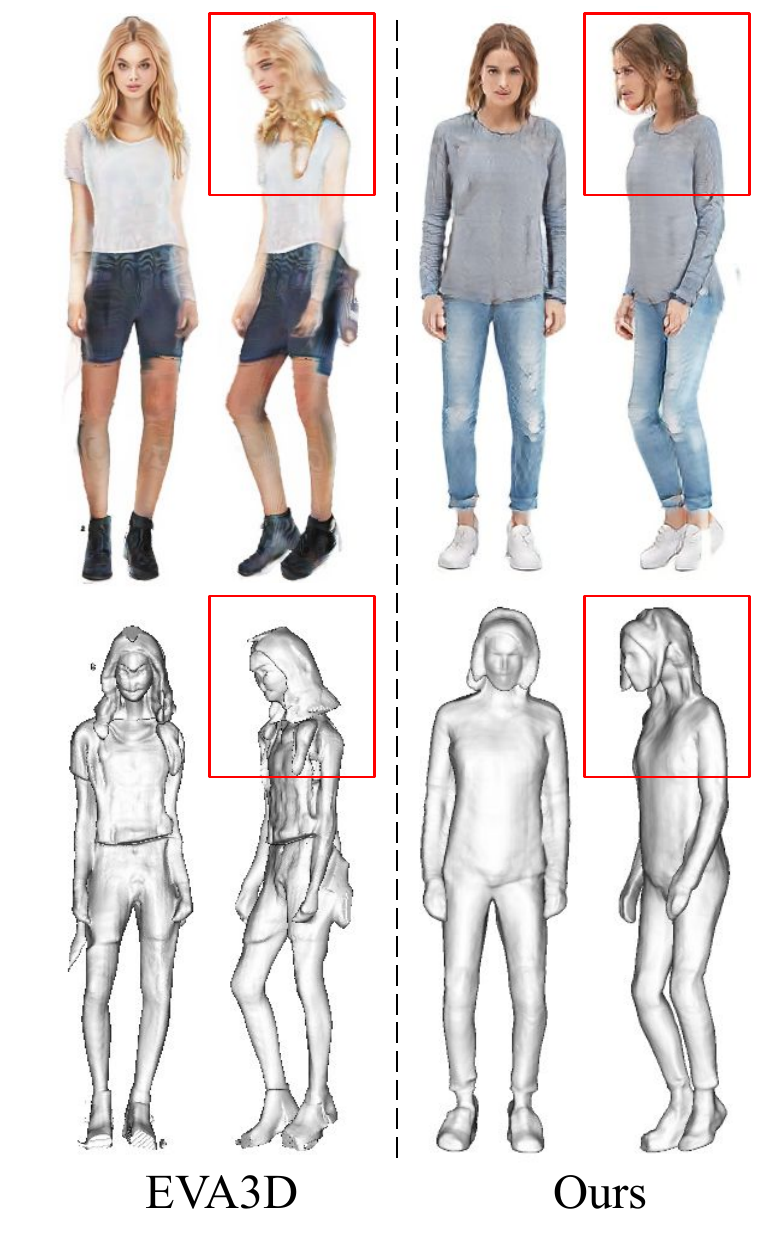}}
         \caption{Novel Views}
         \label{fig:result_compare:view}
     \end{subfigure}
     \hfill
     \begin{subfigure}{0.256\textwidth}
         \centering
         \frame{\includegraphics[width=\textwidth]{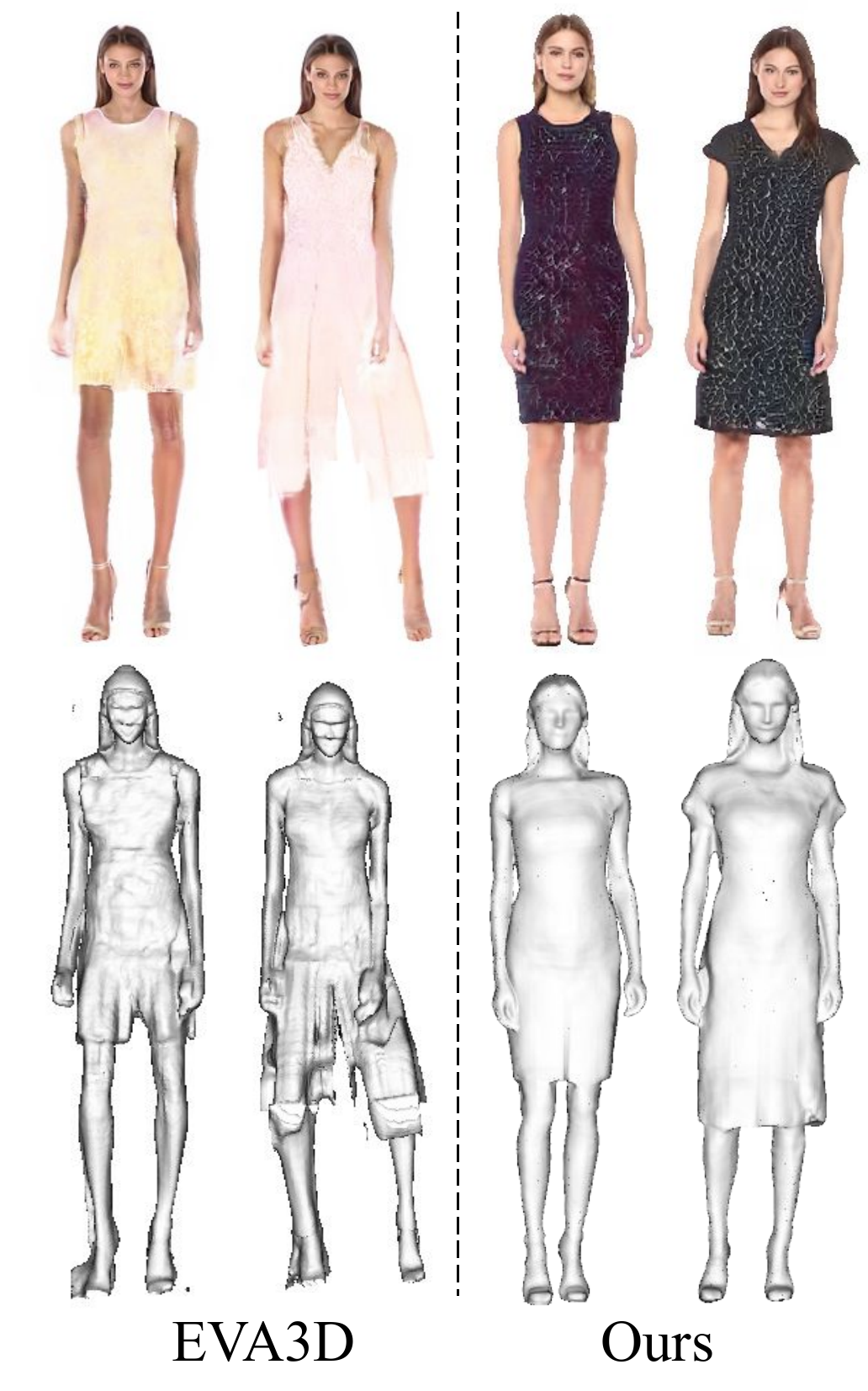}}
         \caption{Loose Clothing}
         \label{fig:result_compare:dress}
     \end{subfigure}
     \vspace{-0.5em}
     \caption{\textbf{Qualitative Comparison to EVA3D.} We show random samples of our method and the \sota method EVA3D~\cite{hong2022eva3d}. Our method achieves better image and shape quality, degrades more gracefully at side views, and better models loose clothing. 
    }
\label{fig:result_compare}
\vspace{-0.3cm}
\end{figure*}

\boldparagraph{Interpolation} Our method learns a smooth latent space of 3D human shape and appearance. As shown in \figref{fig:result_sample:interp}, 
 our method yields smooth transitions of appearance and shape even when interpolating the latent codes of two subjects with different gender and clothing styles.

\subsection{Comparison to \sota}
\label{sec:compare}
\tabref{table:main} summarizes our quantitative comparisons on both DeepFashion and UBCFashion datasets. Since the \sota method EVA3D~\cite{hong2022eva3d} outperforms other baselines by a significant margin, we focus our discussion on the comparison with EVA3D only. More comparisons with other baselines can be found in the \supp.

\boldparagraph{Image Quality} Our method achieves better quantitative results than  EVA3D in terms of  $\fidi$ and $\fidf$ on both datasets. This improvement is confirmed by our user study in \figref{fig:user_study}. Notably, in $81.4\%$ of the cases, participants consider our generated images to be more realistic than EVA3D. A qualitative comparison is shown in \figref{fig:result_compare:overall}.  Our method  generates overall sharper images with more details due to the use of a holistic 3D generator, and synthesizes more realistic faces due to our face discriminators.

Our improvements are particularly pronounced when considering side views. As shown in \figref{fig:result_compare:view}, our method generates sharp and meaningful images also from the side where EVA3D's image quality significantly degrades. 
This is a consequence of EVA3D's pose-guided sampling strategy. As discussed in their paper, EVA3D had to increase the dataset's frontal bias during training to achieve reasonable geometry and face quality. We hypothesize that this requirement is due to the limited capacity of the lightweight part models. As a consequence, EVA3D overfits more to frontal views and generalizes less well. 
In contrast, our efficient articulation and rendering modules allow us to exploit a single holistic generator and our face and normal discriminators enable us to directly sample from the data distribution which leads to better generalization.

Interestingly, the (non-animatable) generative model EG3D achieves reasonable FID despite not modeling articulation. This is because the FID evaluation only considers training poses and views, see also \supp.
\begin{table}
\centering
\resizebox{0.95\linewidth}{!}{ %
\begin{tabular}{@{}lccc@{}}
\toprule
Method & FID $\downarrow$ & $\text{FID}_\text{normal}\downarrow$ & $\text{FID}_\text{face}\downarrow$\\ %
\midrule
Ours & \textbf{10.93} & \textbf{20.38} & 14.79 \\ %
w/o normal GAN & 11.15 & 32.17 & \textbf{14.35}\\ %
w/o face GAN & 11.71 & 23.96 & 20.88 \\ %
\bottomrule
\end{tabular}
} %
\caption{
\textbf{Ablation.} We compare our method and ablated baselines in which we remove individual discriminators.}
\label{table:ablations}
\end{table}

\begin{figure}
    \centering
    \includegraphics[width=\linewidth]{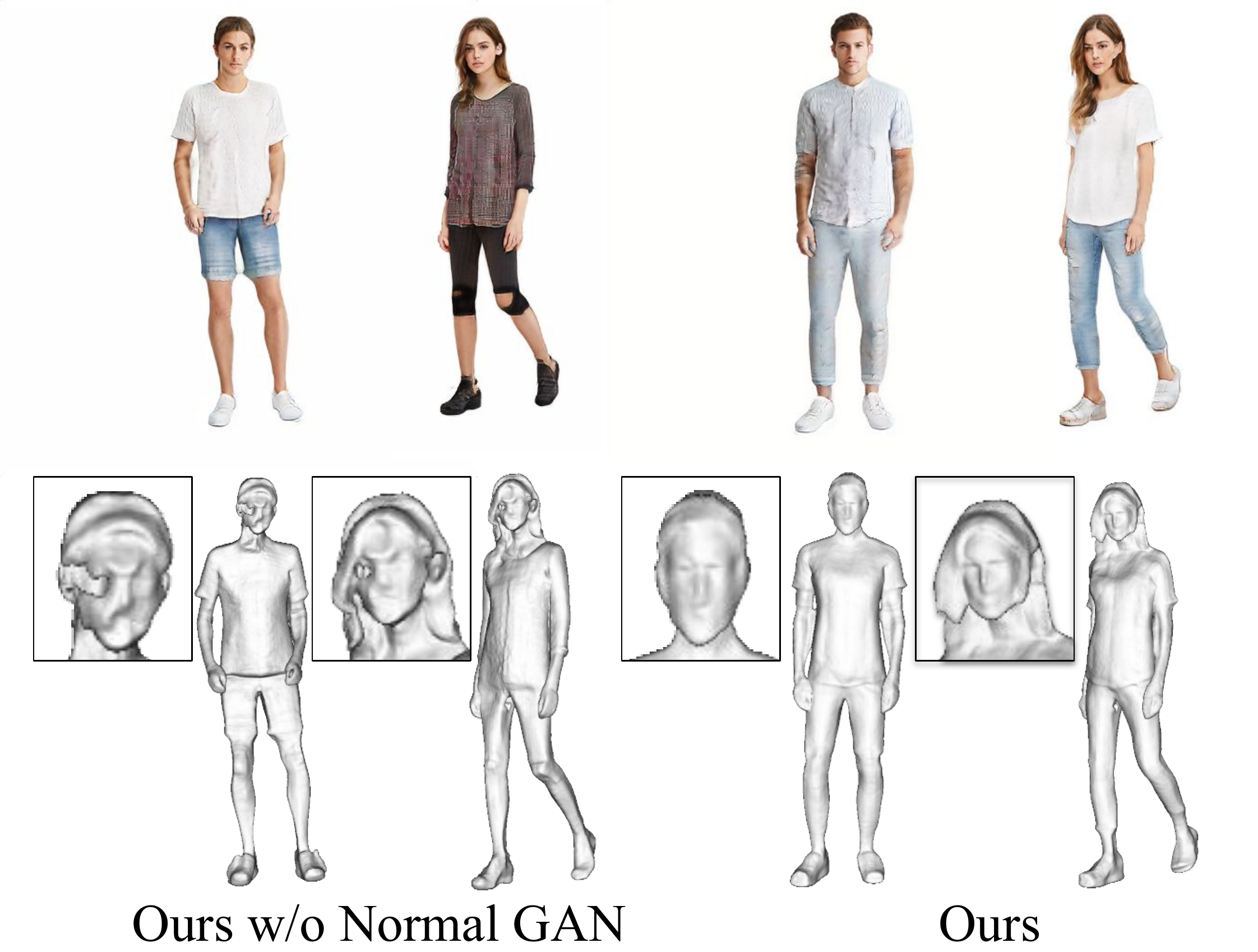}
    \caption{\textbf{Ablation of the Normal Discriminator.} Our normal discriminator effectively improves the generated geometry while preserving appearance quality.}
    \label{fig:result_normal}
    \vspace{-0.3cm}
\end{figure}

\begin{figure}
    \centering
    \includegraphics[width=\linewidth]{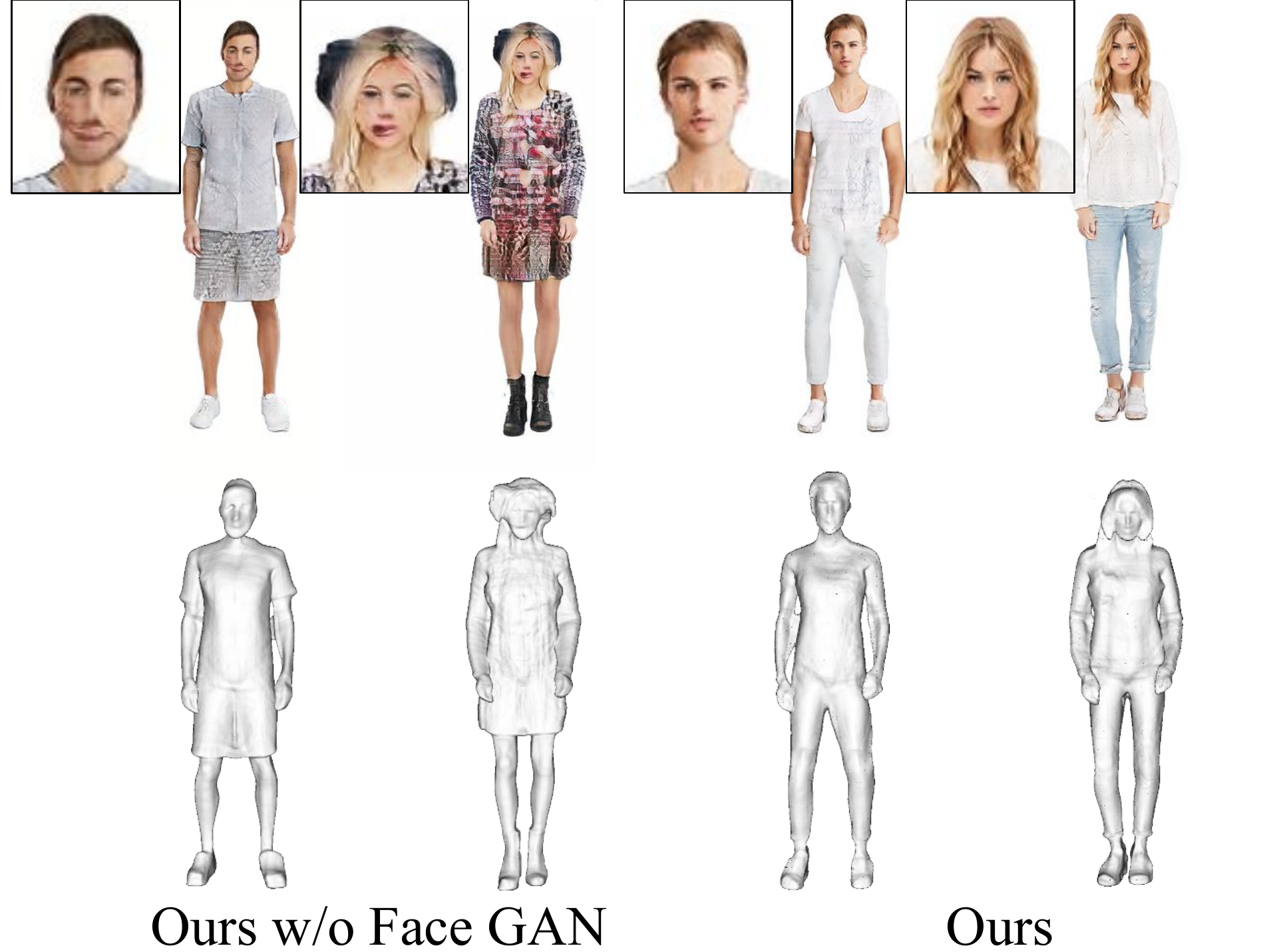}
    \caption{\textbf{Ablation of the Face Discriminator.} Our face discriminator effectively improves generated face quality.}
    \label{fig:result_face}
\end{figure}
\begin{figure}
    \centering
    \includegraphics[width=\linewidth]{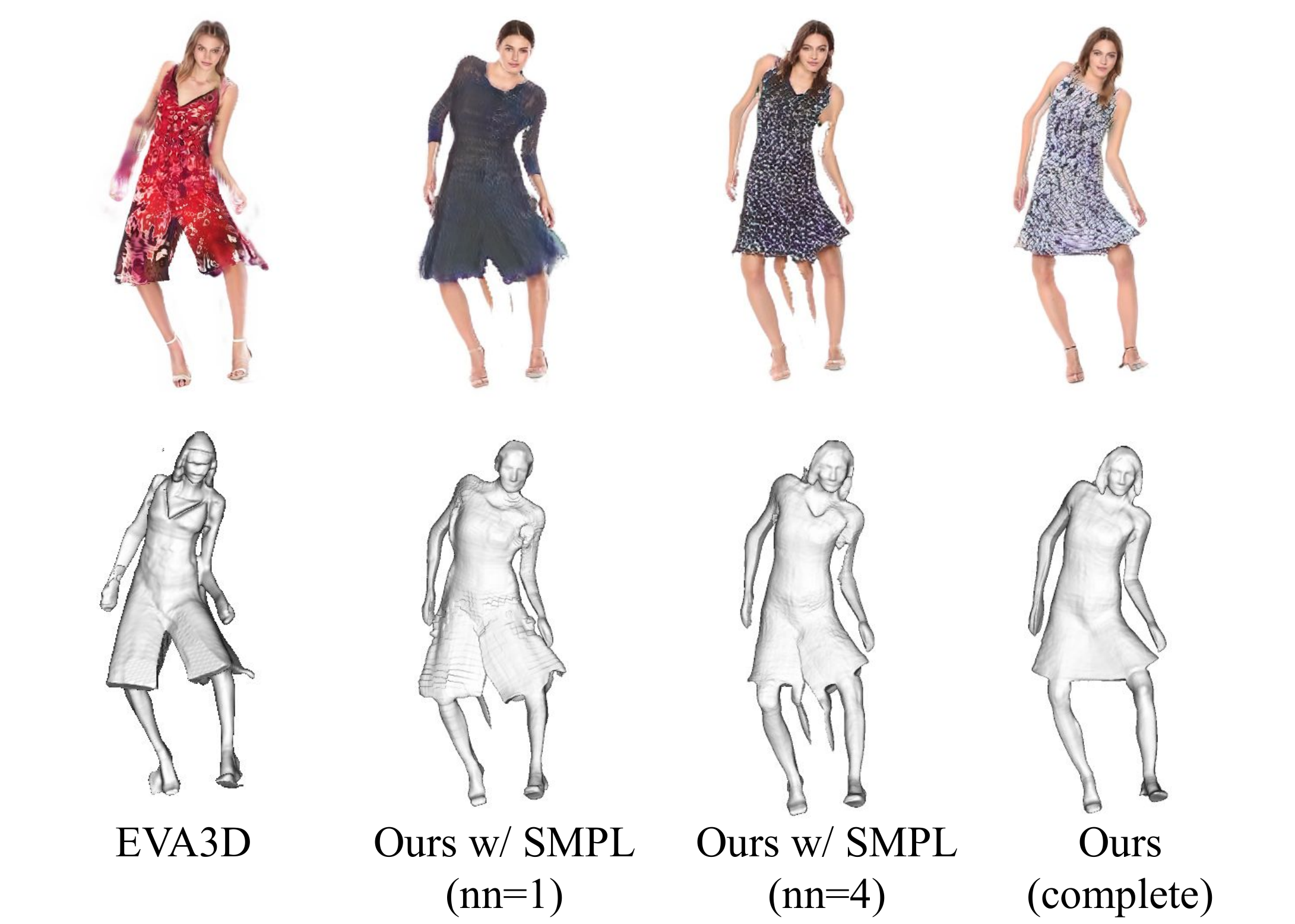}
    \caption{\textbf{Ablation Study of the Deformer.} Results with loose clothing in novel poses, generated by EVA3D and our method with different choices for the articulation module.}     
    \label{fig:result_deform}
    \vspace{-0.3cm}
\end{figure}

\boldparagraph{Geometry} Our method yields significantly better geometry compared to EVA3D, as evidenced by the improvement in $\fidn$ in \tabref{table:main} and the perceptual study in \figref{fig:user_study}. Based on our qualitative results, 
our geometry is more realistic and detailed, in particular on faces.  In contrast, noise and holes can be observed around the shapes generated by EVA3D, despite their surface representation and regularization. We attribute this improvement to our normal discriminators, which provide strong geometric cues.

\boldparagraph{Loose Clothing} As shown in \figref{fig:result_compare:dress}, our method outperforms EVA3D in modeling loose clothing. Due to its compositional nature, EVA3D is prone to generating artifacts between the legs. In contrast, our holistic representation generates loose clothing without discontinuity artifacts. 
\xc{shall we merge this with deformer ablation study?}

\boldparagraph{Efficiency} Our method is more efficient than EVA3D in rendering images and normal maps with the same image and ray sampling resolution. For an image resolution of $256^2$ and with 28 sample points per ray, our method renders normals and images together at 10.5 FPS while EVA3D runs at 5.5 FPS. With a 2D super-resolution module, our method is more than three times faster than EVA3D when rendering images of $512^2$ (9.5FPS vs 3FPS), while achieving better performance in terms of geometry and appearance. More details can be found in \supp.

\subsection{Ablation Study}

\boldparagraph{Normal Discriminator}  
Our normal discriminator serves an important role in improving the realism of the generated geometry.  Comparing our model to ablated versions where we remove the normal discriminator (w/o normal GAN), we observe a significant $\fidn$ improvement (see \tabref{table:ablations}). As shown by the qualitative results in \figref{fig:result_normal}, the normal GAN effectively removes holes and noise on the generated surface, especially on faces, while preserving image quality.

\boldparagraph{Face Discriminator} 
 As expected, when training without face discriminators, we observe a large drop in $\fidf$ (see \tabref{table:ablations}).   Given that faces are low resolution and hard to generate, our adversarial loss on faces forces the generator to focus on this local region and thus achieves a more realistic generation as shown in \figref{fig:result_face}.  

\boldparagraph{Deformer} 
To test the importance of our Fast-SNARF based deformation module, we compare our model to a SMPL nearest-neighbor-based deformer (denoted by \emph{Ours w/ SMPL}), where points are deformed based on the skinning weights of their $K$ nearest SMPL vertices in posed space. As shown in \figref{fig:result_deform}, only our method can deform the skirts without splitting them. This is due to our articulation module being able to derive meaningful deformations for points far away from the SMPL surface. In contrast, our ablated baselines, with different choices of $K$, suffer from discontinuity artifacts as they only provide meaningful deformation at regions close to the SMPL surface. Similar artifacts can be observed in EVA3D's results, which we hypothesize stem from their part-based model.
\subsection{Limitations}
\label{sec:limitations}

Since each training instance is observed only in one pose, the association of pixels to body parts cannot be uniquely determined. Hence, our model sometimes generates wrong clothing patterns under arms, or at hands close to the torso. Future work should investigate techniques to guide association, such as 2D correspondence predictions. \oh{We should probably show this in Supp Mat to illustrate what we mean by this}

Moreover, samples from generative models reflect the biases present in their training data.
The 2D image collections that we used for training focus on fashion images and lack diversity in skin tone, body shape, and age.  
Our work should be viewed as a methodological proof of concept and contains no mechanisms to combat these biases.
To avoid biases future research and deployable systems should i) be trained on more diverse data or ii) use explicit de-biasing.
Further limitations are discussed in the \supp.

\section{Conclusion}
In this paper we contribute a new controllable generative 3D human model that is learned from unstructured 2D image collections alone and does not leverage any 3D supervision. Our model synthesizes high-quality 3D avatars with fine geometric details and models loose clothing more naturally than prior work. 
We achieve this through a new generator design that combines a holistic 3D generator with an efficient and flexible articulation module. Furthermore, we show that employing several, specialized discriminators that operate on the different branches (RGB and normals) and regions (fully body and facial region), leads to higher visual fidelity. 
We experimentally demonstrate that our method advances state-of-the-art in learning 3D human generation from 2D image collections in terms of both appearance and geometry and that it is the first generative model of 3D humans that can handle the deformations of free-flowing and loose garments and long hair.
\vspace{-0.2em}
\boldparagraph{Acknowledgements} Zijian Dong was supported by the BMWi in the project KI Delta Learning (project number 19A19013O) and the ERC Starting Grant LEGO-3D (850533). Andreas Geiger is a member of the Machine Learning Cluster of Excellence, funded by the Deutsche Forschungsgemeinschaft (DFG, German Research Foundation) under Germany’s Excellence Strategy – EXC number 2064/1 – Project number 390727645. Xu Chen was supported by the Max Planck ETH Center for Learning Systems. This project was supported by the ERC Starting Grant LEGO-3D (850533), the BMWi project KI Delta Learning (project number 19A19013O) and the DFG EXC number 2064/1 - project number 390727645. We thank Kashyap Chitta, Katja Schwarz, Takeru Miyato and Seyedmorteza Sadat for their feedback, and Tsvetelina Alexiadis for her help with the user study. 
\noindent\textbf{Disclosure:} 
MJB has received research gift funds from Adobe, Intel, Nvidia, Meta/Facebook, and Amazon.  MJB has financial interests in Amazon, Datagen Technologies, and Meshcapade GmbH.  MJB's research was performed solely at, and funded solely by, the Max Planck.

{\small
\bibliographystyle{ieee_fullname}
\bibliography{egbib}
}

\end{document}